\documentclass[conference]{IEEEtran}
\IEEEoverridecommandlockouts
\usepackage{cite}
\usepackage{amsmath,amssymb,amsfonts}
\usepackage{algorithmic}
\usepackage{graphicx}
\usepackage{textcomp}
\usepackage{blindtext}
\usepackage{float} 
\usepackage{subfigure}
\usepackage{xcolor}
\usepackage{wrapfig}
\usepackage{makecell}

\def\BibTeX{{\rm B\kern-.05em{\sc i\kern-.025em b}\kern-.08em
		T\kern-.1667em\lower.7ex\hbox{E}\kern-.125emX}}
\makeatletter

\begin{document}
	\title{Mobility prediction Based on Machine Learning Algorithms}
	\IEEEpeerreviewmaketitle
	\author{\IEEEauthorblockN{Donglin Wang, Qiuheng Zhou*, Sanket Partani, Anjie Qiu and Hans D. Schotten}
		\IEEEauthorblockA{\textit{University of Kaiserslautern} \\
			Kaiserslautern, Germany \\
			\textit{*German Research Center for Artificial Intelligence(DFKI)} \\
			Kaiserslautern, Germany \\
			$\{$dwang,Partani,qiu,schotten$\}$@eit.uni-kl.de\\
		    $\{$qiuheng.zhou$\}$@dfki.de}
		
	}
	
	\maketitle
\begin{abstract}
Nowadays mobile communication is growing fast in the 5G communication industry. With the increasing capacity requirements and requirements for quality of experience, mobility prediction has been widely applied to mobile communication and has becoming one of the key enablers that utilizes historical traffic information to predict future locations of traffic users, Since accurate mobility prediction can help enable efficient radio resource management, assist route planning, guide vehicle dispatching, or mitigate traffic congestion. However, mobility prediction is a challenging problem due to the complicated traffic network. In the past few years, plenty of researches have been done in this area, including Non-Machine-Learning (Non-ML)-based and Machine-Learning (ML)-based mobility prediction. In this paper, firstly we introduce the state of the art technologies for mobility prediction. Then, we selected Support Vector Machine (SVM) algorithm, the ML algorithm for practical traffic date training. Lastly, we analyse the simulation results for mobility prediction and introduce a future work plan where mobility prediction will be applied for improving mobile communication. 
\end{abstract}

\begin{IEEEkeywords}
Mobility prediction, ML, SVM
\end{IEEEkeywords}

\section{Introduction}
With the development of modern society, more applications of mobile communication are required to operate in highly dynamic pervasive computing environments on the road network. In order to meet the requirements, Predicting the location of mobile Users (UEs) e.p., traffic vehicles or pedestrians is used which can help those traffic participates act proactively. Also, accurate mobility prediction can help route planning, vehicle dispatching, mitigate traffic congestion or reduce handover frequency, etc. So in this paper, we focus on the traffic user mobility location prediction that plays an essential role in the intelligent transportation systems. Recently, there is a significant amount of state of the art researches completed the mobility prediction, technically these works are divided into Non-ML-based user mobility prediction and others are ML-based user mobility prediction. 

In [1], Lagrange interpolation is utilized, and it is Non-ML-based. An UE with a selected destination and randomly selected speed is used to describe the position and speed of the UE. Based on the traffic information of the UE, the Lagrange interpolation in Descartes coordinate system is used here to simulate movement history information of UE. The UE position coordinates are stored as $(x_i,y_i)$, where $i=0,1,2…n$. By using Lagrange interpolation formula, the predicted position and direction of UE are generated based on the UE trajectory. Combining the predicted position $(x_p,y_p)$ and direction, we can have a completed construction of the UE mobility model. [2] proposes Long Short-Term Memory (LSTM) networks and Dead Reckoning (DR) framework which is a Non-ML-based approach for predicting mobility. In this work, a proactive mobility management solution based on virtual cells is proposed, motivated trajectory prediction using a real-world vehicle mobility dataset. The proposed LSTM networks and DR framework predicts the next locations of vehicles by integrating the LSTM-DR method. Simulation results show that higher accuracy and robustness in trajectory prediction can be achieved by the proposed prediction framework. 

Nowadays, many ML methods are introduced and developed for different fields, which can produce excellent outcomes and show distinct advantages with large datasets. For traffic mobility prediction, there are some particular ML methods used, such as Deep Neural Network (DNN), Extreme Gradient Boosting Trees (XGBoost), Semi-Markov, and SVM. 

The research [3] analysed DNN, XGBoost, Semi-Markov, and SVM found in fig.1. In this investigation, the realistic synthetic dataset of mobile users through a realistic Self-similar Least Action Walk mobility model is generated and applied. The performance of each ML model based on the mobility model is evaluated to not only predict the future location of mobile users but also the time each algorithm takes to be fully trained. 
\begin{figure}[htbp]
	\centering
	\includegraphics[width=\linewidth]{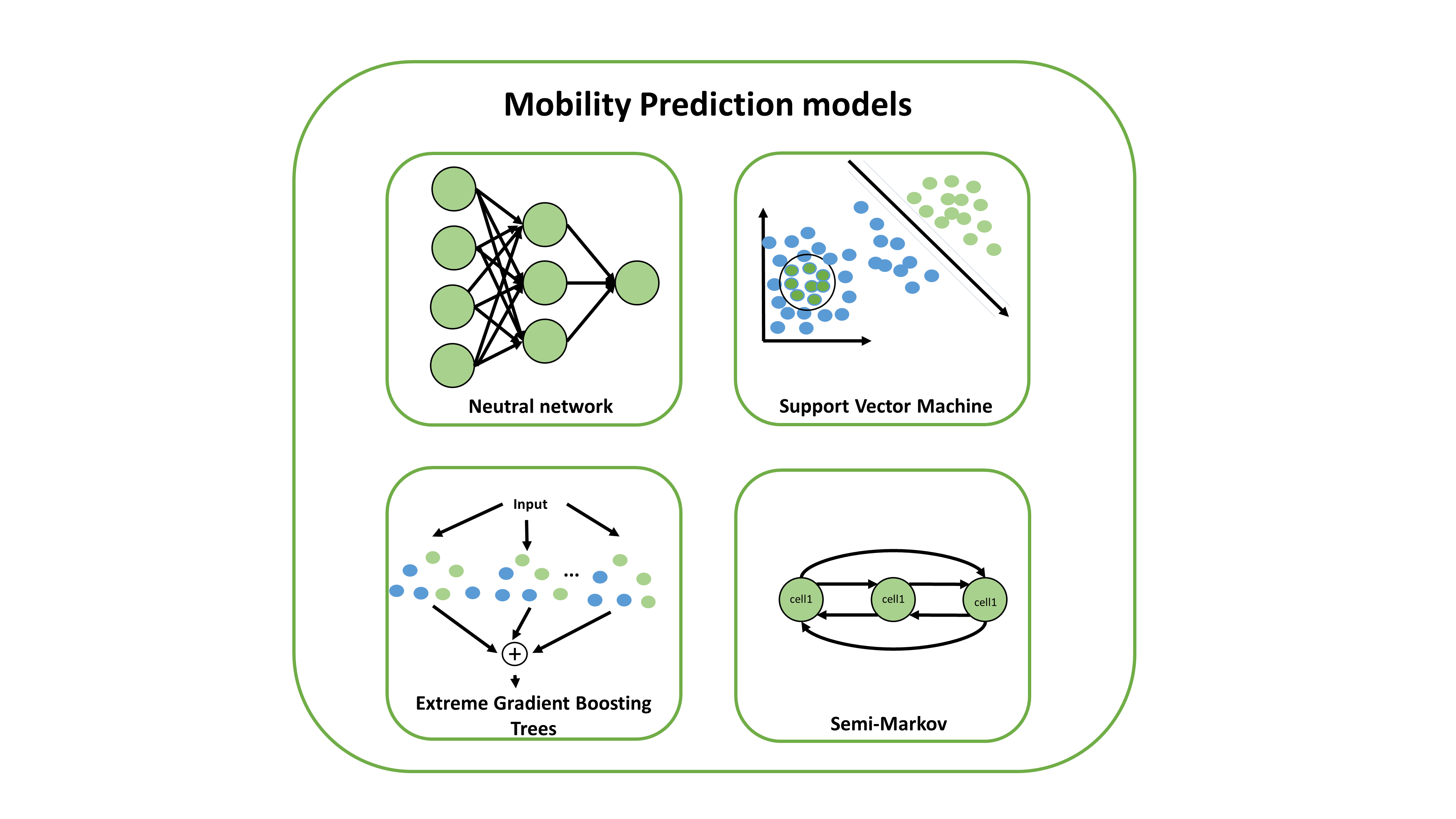}
	\caption{Mobility prediction models}
	\label{fig}
\end{figure}

In this work, the benefits of each ML algorithm are generated and analysed to meet different requirements. Its contribution is to qualify the performance of each algorithm to find the best solution for mobility prediction in mobile networks. According to the simulation results, it shows XGBoost is an effective mobility prediction algorithm and the predictive accuracy of this algorithm is relatively high with a lower execution time. The SVM gives a high accuracy rate compared to the DNN and semi-Markov algorithms [3]. [4],[5] and [6] are all ML-based researches for mobility prediction. In the paper [4], general mobility prediction based on the ML model is introduced. The location information considered for classification refers to the history of user movements location information pre-evaluation of UE can be matched with information classification or prediction, in the way that, the values of certain parameters for location estimation are determined in advance. A location model is proposed in order to support location prediction for mobile users. Experiments with the Voting classification algorithm which is implemented in the Weka ML workbench are performed. Previous work including Kalman Filter-based predictor and Hidden-Markov model approach for the location prediction, pattern-matching algorithm – Hierarchical Location Prediction (HLP) are considered. In [5], the SVM algorithm is used to predict the vehicular user route. Supervised learning takes input variables $(X)$ and output variables $(Y)$ and uses an algorithm to learn the mapping function from the input to the output $Y = f(X)$. The goal of the model training is to train a prediction model that can find the relationship between the prediction values $y_i$ (route to be taken by the user) and instance $x_i(u_i,t_i)$ by using the historical route information of the users $(x_1,y_1),...(x_n,y_n)$. When a new value $x_k(u_k,t_k)$ is given to the model, the model can predict the next route $y_k$, the user $u_k$ will be taking at that particular time slot $t_k$. In [6], an advanced Multi-class Support Vector Machine Based Mobility Prediction (Multi-SVMMP) scheme to estimate the future location of mobile users according to the trajectory of each user is introduced. The regular and random user movement patterns are treated differently in the location prediction process, which can reflect the user movements more realistically than the existing movement models. And different forms of multi-class support vector machines are embedded in the two mobility patterns according to the different characteristics of the two mobility patterns [6]. As the simulation results turn, this Multi-SVMMP can solve the traditional problem, obtain a higher prediction accuracy, improve user adaptability, and reduce the cost of prediction resources.

\section{SVM model}
As SVM is one of the supervised ML methods, which is the most widely used for UE's mobility prediction. In this chapter, a brief introduction is made to SVM algorithm. Further, a practical mobile dataset will be generated and the SVM algorithm will be applied to train the dataset.  
SVM is a supervised linear ML model that uses classification algorithms for group classification and regression problems. Linear and non-linear problems and many practical problems can be solved nicely by applying the SVM algorithm. Also, compared to other algorithms like NN, SVM has higher training speed and better performance with a limited number of samples as mentioned in the former context. This makes the algorithm very suitable for text classification problems as well. So SVM is the appropriate way to be used for mobility prediction where provides lots of traffic text information. 
Following is the simple idea of SVM: Firstly, the SVM algorithm creates a line or a hyperplane (so-called decision boundary) in N-dimensional space (N- the number of features) which distinctly separates the data into classes. There are plenty of possible methods to choose hyperplanes used for deciding the classes of the datasets. 
Secondly, we need to define a margin. The objective of data training is to find a plane that has the maximum margin i.e., the maximum distance between the data points of all the classes which are based on what our model can learn from the training dataset. By maximizing the size of this margin, we can build an optimal classifier as shown in fig.2.

\begin{figure}[htbp]
	\centering
	\includegraphics[width=\linewidth]{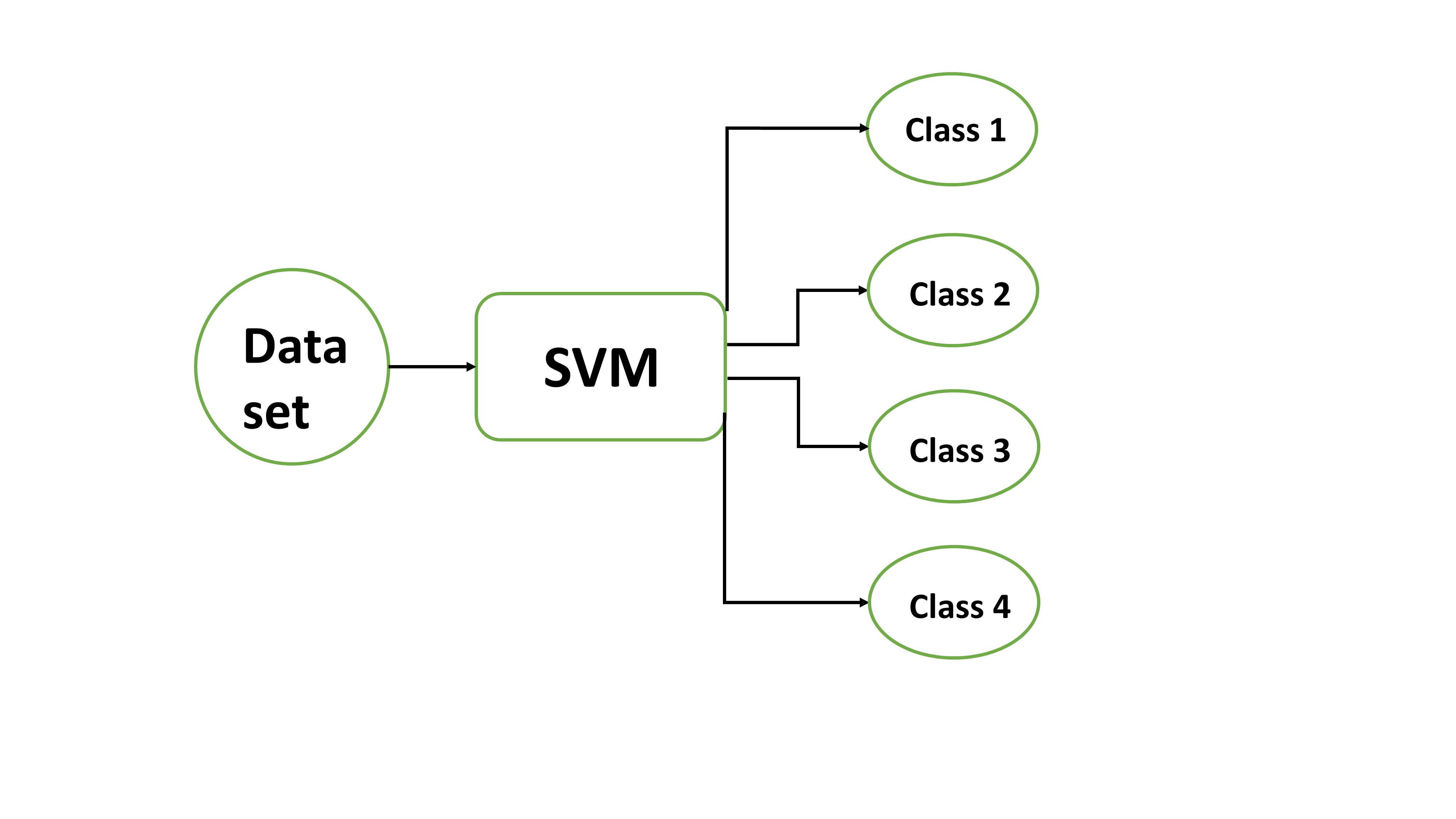}
	\caption{SVM classifier }
	\label{fig}
\end{figure}

Here, we have the standard way of writing our model for SVM model is 
\begin{equation}
z(x)=(w^Tx+b).
\end{equation}
More completely,the classification function, $g(z)$ is defined as:
\begin{equation}
g(z)= 
\begin{cases}
1,              & \text{if} z\geq 0 \\
-1,              & \text{if} z\leq 0,
\end{cases}
\end{equation}
so combined with $z(x)$,the complete model for SVM is:
\begin{equation}
g(z(x))=g(w^Tx+b).
\end{equation}
In fig.3, we have a practical model with two classes "+" and "-". As shown in the figure, several possible hyperplanes A, B, and C have be defined to separate the two classes. Our goal is to construct a hyperplane that is consistent with the data. The most possible space (margin) between the decision boundary and the data points on each side of the line in order to increase the total prediction confidence. An optimal classifier is constructed by maximizing the size of this margin. Practically, the functional margin and the geometric margin are two ways used for measuring the margin [7]. 
For simplicity,  definitions for these two margin calculation ways will be explained shortly here.
\begin{figure}[htbp]
	\centering
	\includegraphics[width=\linewidth]{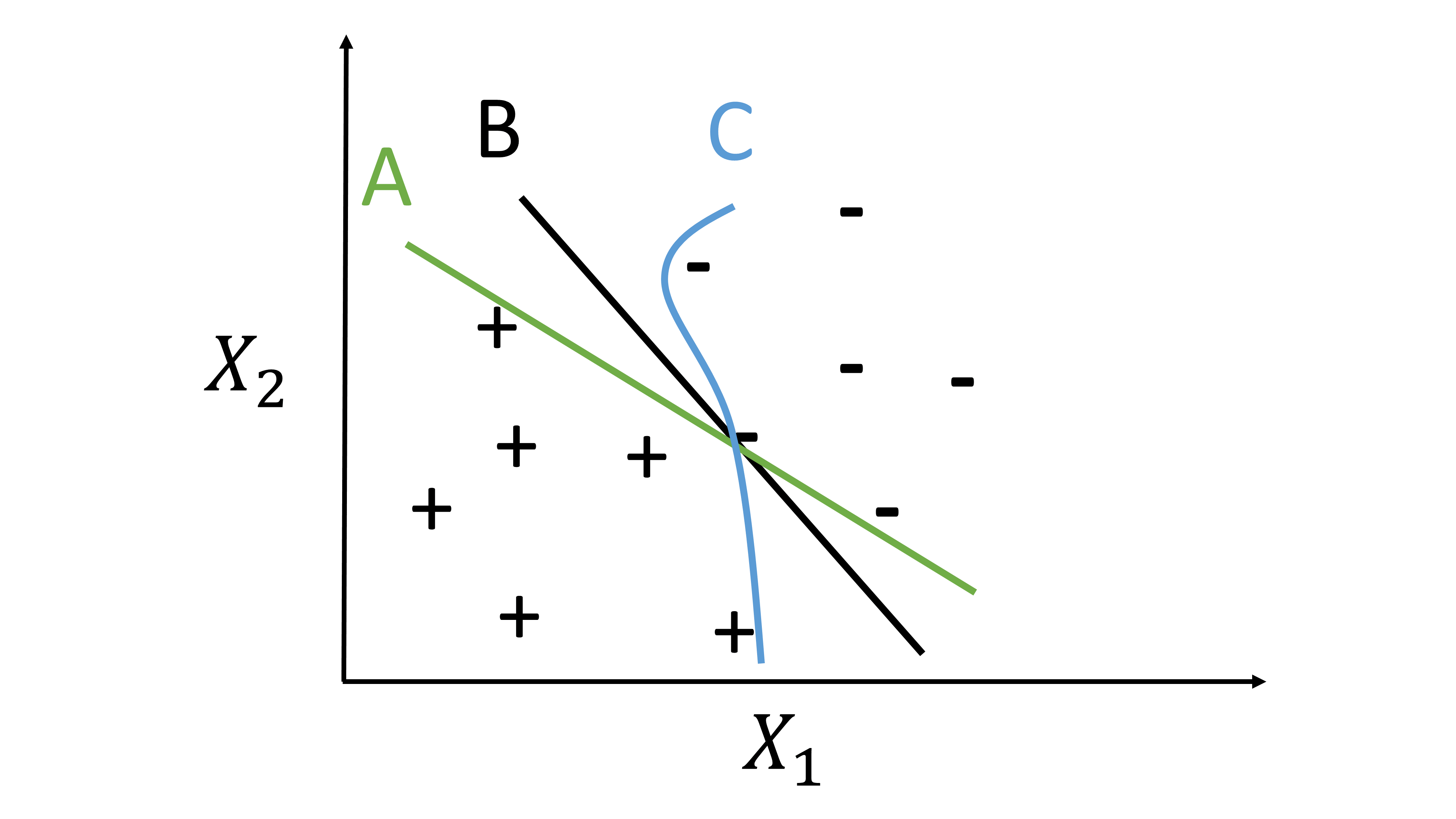}
	\caption{Possible hyperplanes}
	\label{fig}
\end{figure} 
\begin{itemize}
	\item functional margin definition
\end{itemize}
The functional margin is defined as:
\begin{equation}
\hat{\gamma}_i=y_i(w^Tx+b),
\end{equation}
where $y_i$ is the actual class, and $(w^Tx+b)$ is the classifier's prediction. The margin is calculated by comparing the classifier prediction to the actual class. 
From the equation, we can find if $y_i=1$, a large position number for the rest $w^Tx+b$ is needed due to a large functional margin that we want to have. Vice versa, when the $y_i=0$, a large negative number would be required.
In order to reach our goal for a maximum boundary, the value of $w$ and $b$ are arbitrarily scaled. 

\begin{itemize}
	\item geometric margin definition 
\end{itemize}
The definition of geometric margin is: 
\begin{equation}
{\gamma}_i=y_i((\frac{w}{\mid w \mid})^T x_i+\frac {b}{\mid w \mid} ),
\end{equation}
where $\frac{w}{\mid w \mid}$ represents the unit direction normal to the decision boundary. 
The distance from a data point to the decision boundary is used to define the geometric margin. 
To get more details for functional margin and the geometric margin, you can check [7] where clearly provides all the definitions and calculation processes. 

After defining the space between the data and hyperplane, the next problem is to maximize the space. The straightforward approach towards finding the parameters for a hyperplane with the most space between the data and the corresponding decision boundary is to simply maximize the geometric margin. There are lots of ways for optimizing the parameters in order to get the maximum space. An import kernel function has been created, moreover,  common kernel functions include: linear, polynomial, rbf, sigmoid which will be used in the simulation.
Finally, after giving an SVM model sets of labelled training data for each group using the features provided, i.e. every example on the same of decision boundary has the same label. They’re able to categorize new data.
Above are the most simple processes of SVM algorithm, for more information regarding SVM are in [7][8][9]. 
 
\subsection{System mode}
We deploy a highway scenario with two routes, three lanes, and one direction as illustrated in fig.4 with SUMO [10]. Random driving cars are generated and running on this highway by using SUMO. At the junction point, some of these vehicles drive straightly which means these vehicles choose Route1 and are marked with the number "0". Some of the others choose to turn right which means those cars are taking Route2 and marked with a number "1". With these "0" and "1" two marks, vehicles are easily divided into two classes. We let these vehicles run on the highway for 100 time steps and get the context of each vehicular mobility e.g., traffic ID, the longitude and latitude of mobility, and traffic mark,ect. So the origin train mobility dataset for SVM algorithm is generated and will be used in the SVM algorithm simulation.   

\begin{figure}[htbp]
	\centering
	\includegraphics[width=\linewidth]{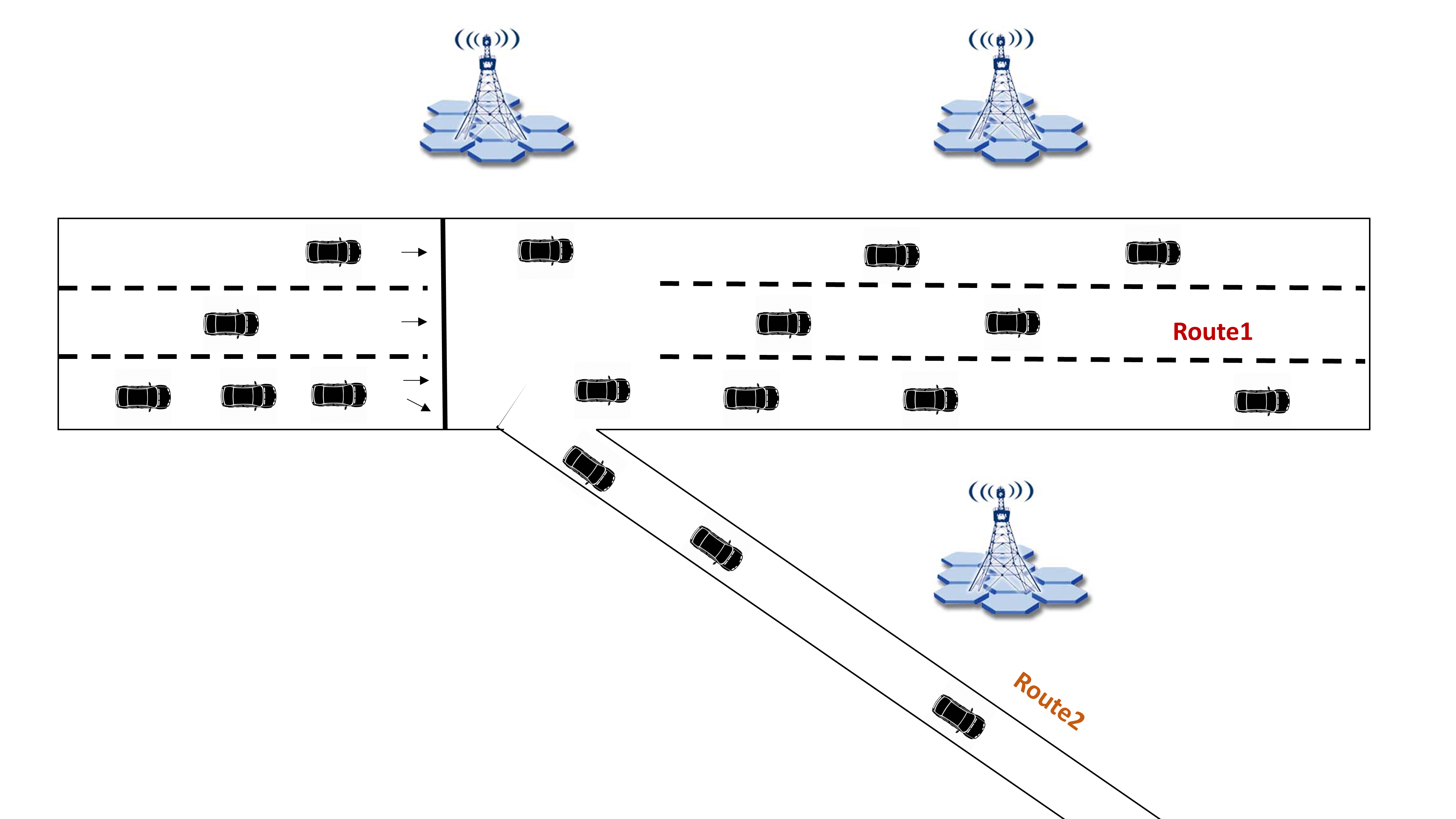}
	\caption{Highway scenario}
	\label{fig}
\end{figure}

\begin{figure}[htbp]
	\centering
	\includegraphics[width=\linewidth]{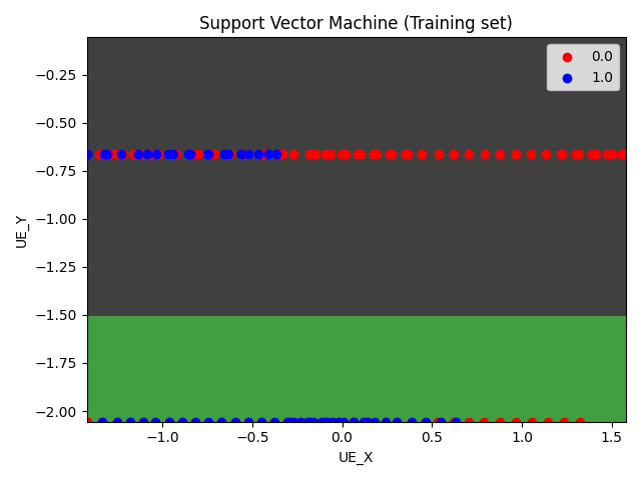}
	\caption{SVM training set for 400 examples}
	\label{fig}
\end{figure}

\begin{figure*}[htbp]
	\centering
	\subfigure[SVM testing 10 examples]{
		\includegraphics[width=0.45\textwidth,height=0.7\columnwidth]{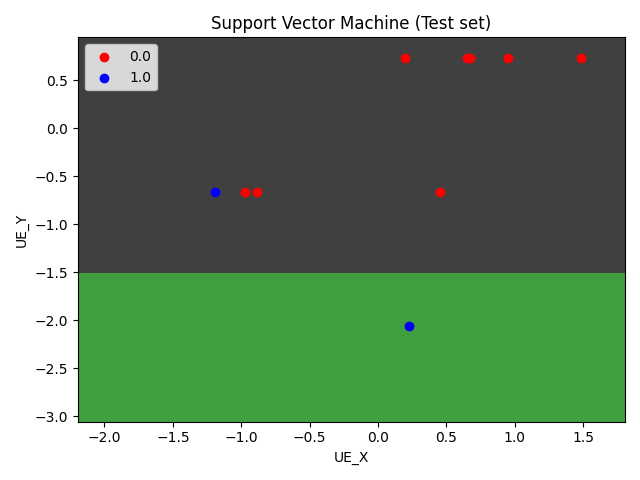}}
	\label{fig.sub.1}
	\subfigure[SVM testing 100 examples]{
		\includegraphics[width=0.45\textwidth,height=0.7\columnwidth]{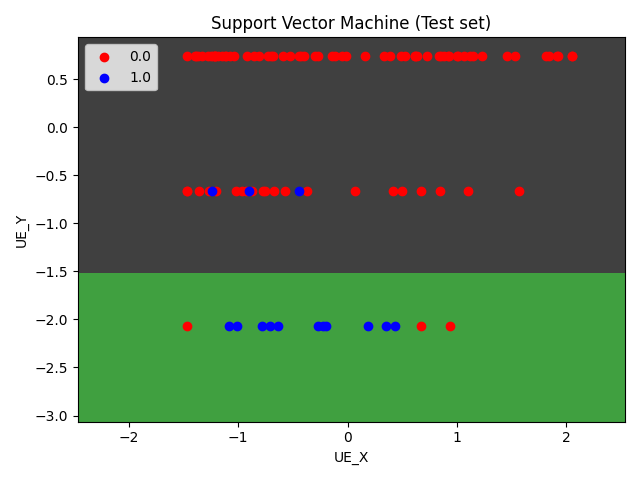}}
	\label{fig.sub.2}
	\caption{The effect on prediction accuracy of different testing data}
	\label{fig}
\end{figure*}

\subsection{Python library}
There are already lots of python libraries where implemented the SVM algorithm in all different ways. In this simulation, we use a simple implementation of an SVM model in python. The classifier is an object of the SVM class which was imported from sklearn.svm library [11]. Also, the linear kernel type is chosen.

\section{training ans testing results}
Since we have the simulator in python for the SVM algorithm, and the origin training dataset has been generated in SUMO. 
The simulation results are shown in the following figures. Fig.5 is illustrating the chosen hyperplane according to the “Y” axis of the traffic user. In fig.5, the training data examples of traffic users are 400. After training these data examples by SVM algorithm, we have the linear decision boundary where is seen as $Y=-1.5$, which means when the traffic mobility is distributed at the dark part above the hyperplane at a time point, the traffic user is predicted to take the Route1, vice versa, if the traffic mobility is driving and located at the green part lower the hyperplane, we assume the traffic user will take the Route2 at the time point. 

In fig.6(a) and fig.6(b), we use 10 test data examples and 100 test data examples separately in order to check the prediction accuracy by using the SVM algorithm for 400 training examples. In fig.6(a), the red dot and blue dot represent "0" and "1" respectively. A blue dot is seen at the dark region which is supposed to be at the lower green part. We take this situation as a false prediction, so the prediction accuracy is 9/10 where 9 mobility predictions are correct and 1 mobility prediction is wrong. When we increase the amount of testing data, more false predictions can be found. There are three blue dots in the dark area and three red dots in the green area that are incorrect predictions. In this case, the mobility prediction accuracy is 94/100. 

\begin{table}[htbp]
	\caption{Mobility prediction accuracy}
	\begin{center}
		\begin{tabular}{|l|l|}
			\hline 
			Testing examples &	400 training examples\\
			\hline
			10&		90\% \\  
			\hline
			20&		90\% \\  
			\hline
			30&		86.07\% \\  
			\hline
			40&		92.5\% \\  
			\hline
			50&		94\% \\  
			\hline
			60&		95\% \\  
			\hline
			70&	    92.8\% \\  
			\hline
			80&     93.75\% \\  
			\hline
			90&	    93.3\% \\  
			\hline
			100&    94\% 	\\  
			\hline
		\end{tabular}
	\end{center}
\end{table}

In Tab.I, more values of prediction accuracy of mobility for different testing sizes are given. It's obvious to find most values of prediction accuracy are above 90\% and the average prediction accuracy is 92.2\%, which means the chosen decision boundary for this scenario is quite valuable and can predict the traffic user to take which route. 

\section{conclusion}
According to the SVM algorithm, we can predict which route the traffic user will take, and the prediction accuracies of different training datasets have been generated and to check whether the chosen hyperplane for mobility prediction for a specific scenario is good. Since the simulation results are usable, mobility prediction will advantage for further handover resource allocation, detect traffic congestion conditions, route planning, vehicle dispatching,ect. Future work includes considering applying the mobility prediction for handover process and to improve service migration. Further, more ML algorithms can be used to compare prediction accuracy.A real-world highway scenario is considered and this SVM algorithm will be optimized for training more complex dataset like the urban city situation. 
This paper is an introduction for a reader to get to know there are lots of ML algorithms for mobility prediction, and how to apply the SVM algorithm for practical data training. Moreover, it can make people know the advantages of mobility prediction which can be used extensively in cellular network or for smart traffic cities.

\end{document}